\documentclass[twoside,11pt]{article}

\usepackage[abbrvbib, preprint]{jmlr2e}

\usepackage{microtype}
\usepackage{booktabs} 
\usepackage[utf8]{inputenc}
\usepackage[T1]{fontenc}
\usepackage{url}
\usepackage{booktabs}
\usepackage{amsfonts}
\usepackage{dsfont}
\usepackage{nicefrac}
\usepackage{amsmath}
\usepackage{amssymb}
\usepackage{amsfonts}
\usepackage{graphicx}
\usepackage{subfigure} 
\usepackage{stmaryrd}

\newcommand\ex{\mathbb{E}}

\renewcommand\mathbb{\mathds}
\newcommand\err{\mathrm{err}}

\ShortHeadings{On Localized Discrepancy for Domain Adaptation}{Zhang, Long, Wang and Jordan}
\firstpageno{1}

\begin{document}

\title{On Localized Discrepancy for Domain Adaptation}

\author{\name Yuchen Zhang
	\email yuczhang17@gmail.com \\
	\addr School of Software and BNRist\\
	Tsinghua University\\
	Beijing 100084, China
	\AND
	\name Mingsheng Long\thanks{Corresponding author: Mingsheng Long (mingsheng@tsinghua.edu.cn)}
	\email mingsheng@tsinghua.edu.cn \\
	\addr School of Software and BNRist\\
	Tsinghua University\\
	Beijing 100084, China
	\AND
	\name Jianmin Wang
	\email jimwang@tsinghua.edu.cn \\
	\addr School of Software and BNRist\\
	Tsinghua University\\
	Beijing 100084, China
	\AND
	\name Michael I.\ Jordan \email jordan@cs.berkeley.edu \\
	\addr Division of Computer Science and Department of Statistics\\
	University of California\\
	Berkeley, CA 94720-1776, USA}


\maketitle

\begin{abstract}
	We propose the discrepancy-based generalization theories for unsupervised domain adaptation. Previous theories introduced distribution discrepancies defined as the supremum over complete hypothesis space. The hypothesis space may contain hypotheses that lead to unnecessary overestimation of the risk bound. This paper studies the localized discrepancies defined on the hypothesis space after localization. First, we show that these discrepancies have desirable properties. They could be significantly smaller than the pervious discrepancies. Their values will be different if we exchange the two domains, thus can reveal asymmetric transfer difficulties. Next, we derive improved generalization bounds with these discrepancies. We show that the discrepancies could influence the rate of the sample complexity. Finally, we further extend the localized discrepancies for achieving super transfer and derive generalization bounds that could be even more sample-efficient on source domain.
\end{abstract}


\section{Introduction}

General learning theories assume that the training and test data are generated by identical underlying distribution. 
This assumption may not hold in many applications. 
Some tasks may need to learn with some available training data to test on distinct distribution without labeled data. 
This problem is known as domain adaptation \cite{cite:Book09DSS} that attracts a great deal of attention recently.
A mainstream theoretical framework for domain adaptation is the discrepancy-based theories. 
Ben-David \textit{et al.} \cite{cite:NIPS07DAT} used a parameterized discrepancy to measure the error gap between domains in binary classification tasks. 
Later, this group of theories were developed from multiple perspectives by a series of works \cite{cite:COLT09DAT, thy:germain2013bayes,thy:cortes2015gd,thy:zhang2019bridging}. They share the desirable property that discrepancies can be estimated from finite unlabeled datasets and provide guidance to algorithm designs for domain adaptation.

A relatively small discrepancy is necessary for the success of domain adaptation.
Classical discrepancies \cite{cite:NIPS07DAT, cite:COLT09DAT} are defined as the supremum over a pair of complete hypothesis spaces.
These definitions are relatively conservative since they include many hypotheses that are impossible to have good performance on either domains.
Such hypotheses could make these discrepancies excessively large.
Several works try to solve this problem by proposing new discrepancies defined on the hypothesis spaces after localization. 
\cite{thy:cortes2015gd} uses a small amount of target labeled data to reduce one of the paired hypothesis spaces.
\cite{thy:zhang2019bridging} replaces one of the paired hypothesis spaces to the main hypothesis. Still,
there are no provable guarantees that the localization on discrepancy brings generalization advantages. 

In our paper, we study the localized discrepancies with an alternative localization method. 
We prove that after localization, these discrepancies can also lead to target error bounds under mild assumptions.
We find that the localized discrepancies endow impressive properties. 
The localized discrepancies could be smaller than previous discrepancies by a large margin. 
They are able to reveal asymmetric transfer difficulties since their values will be different after exchanging the source and target domains.
We propose generalization bounds for these localized discrepancies. 
We show that after localization, the sample complexities are controlled by the discrepancies and could be faster when they are small enough. 
This is the first theoretical result that shows discrepancy-based upper bounds could achieve faster sample complexities under classical assumptions \cite{cite:NIPS07DAT}. 
We also analyze how to achieve super transfer \cite{thy:hanneke2019on} in unsupervised domain adaptation. 
Super transfer refers to generalization bounds where the order of source sample complexity is faster than that of the target.
We prove that the boosted versions of the localized discrepancies can lead to super transfer without excessive assumptions, thus can be even more sample-efficient.

Overall, we summarize our contributions as the following aspects:
\begin{itemize}
	\item We define new localized discrepancies with an alternative localization technique. We prove that these localized discrepancies could lead to transfer error bound and have good properties.  These discrepancies can be strictly smaller than classical discrepancies.  They can also reveal the asymmetric transfer difficulties in domain adaptation.
	\item We propose improved generalization bounds with these localized discrepancies. We prove that the value of these discrepancies could also influence the generalization errors. When the discrepancies are small enough, the sample complexity can be faster.  
	\item We further study super transfer in unsupervised domain adaptation. With boosted versions of the localized discrepancies, we could achieve faster generalization rate on source domain than on target domain, which could thus  be further more sample-efficient. 
\end{itemize} 
Our paper is organized as follows. We first introduce basic notations and settings in Section \ref{sec2}. 
We then define our localized discrepancies in Section \ref{sec3} and further propose upper bounds of target error. We study the properties of localized discrepancies in Section \ref{sec4}. We propose improved generalization bounds with localized discrepancies in Section \ref{sec5}. Finally, we extend the localized discrepancies to enable super transfer in Section \ref{sec6}. Further analysis and all proofs could be found in appendix.

\section{Preliminaries}\label{sec2}

In supervised learning setting, we learn from an original distribution $D$ on $\mathcal X \times \mathcal Y$, where $\mathcal X$ is the instance space and $\mathcal{Y}$ is the output space. Throughout learning, we first sample a dataset with $n$ labeled points $D_n = \{(x_i,y_i)\}_{i=1}^n$ independently and identically distributed (i.i.d.) from $D$. We then define a hypothesis space $\mathcal{H} : \mathcal{X}\to \mathcal{Y}$ and find the best hypothesis $h\in \mathcal{H}$ that makes the fewest mistakes on distribution $D$. A loss function $L$ is used to measure the error of $h$. The \emph{expected error} of hypothesis $h$ on distribution $D$ is defined as $\err_{D}(h) \triangleq \ex_{(x,y)\sim D} L(h(x),y)$. While the original distribution $D$ is unknown, we can estimate the expected error by the \emph{empirical error} of $h$ on dataset $D_n$ as $\err_{D_n} (h) \triangleq \frac{1}{n}\sum_{i=1}^{n} L(h(x_i),y_i)$.

In domain adaptation setting, we learn from two different distributions on $\mathcal X \times \mathcal Y$: source distribution $P$ and target distribution $Q$. We sample a labeled source dataset $P_n = \{(x_i,y_i)\}_{i=1}^n$ and an unlabeled target dataset $Q_m = \{(x_i)\}_{i=n+1}^{n+m}$. Our goal is to find a good hypothesis $h\in\mathcal{H}$ with low expected error $\err_{Q}(h)$ on target distribution $Q$, which is challenging since target dataset $Q_m$ is unlabeled. The basic idea of domain adaptation is to find a good hypothesis $h$ on source distribution $P$ with low expected error $\err_{P}(h)$ and adapt it to the target distribution $Q$, based on datasets $P_n$ and $Q_m$.

In this paper, we mainly consider binary classification tasks, where $\mathcal{Y} = \{0,1\}$. Unless specifically mentioned, the loss $L$ is set as 0-1 loss: $L(y_1, y_2) = \mathbf{1}[y_1\ne y_2]$. This loss takes value in $\{0,1\}$ and is symmetric. Denote by $d$ the VC-dimension \cite{vapnik1995the} for measuring the complexity of hypothesis space. 

\section{Localized Discrepancies}\label{sec3}
\label{sec:theory}

In this section, we propose the definition of localized discrepancies and study their mathematical properties. 
We first need some proper statistics to measure the distribution shift from $P$ to $Q$.
A series of previous works focus on designing distribution discrepancies for domain adaptation. 
Ben-David \textit{et al.} \cite{thy:shai10ad} proposed the first finite-sample estimable statistic for 0-1 loss, the \textit{{$\mathcal H\Delta\mathcal H$-divergence}}:
\begin{equation}
\mathrm{disc}_{\mathcal{H}}(P,Q) = \sup_{h, h'\in \mathcal H} | \mathbb{E}_{ Q} L(h',h) - \mathbb{E}_{ P}L(h',h)|.
\end{equation}
This $\mathcal{H}$-dependent divergence is defined as the supremum on the symmetric difference hypothesis space $\mathcal H\Delta\mathcal H=\{L(h',h) | h, h'\in \mathcal H\}$.
It was generalized by Mansour \textit{et al.} \cite{thy:mohri2009dd} to the \textit{discrepancy distance} for a broader class of loss functions.
Recently, Zhang \textit{et al.} \cite{thy:zhang2019bridging} introduced the \textit{disparity discrepancy} which takes the supremum on a hypothesis-dependent space $\{L(h',h) | h' \in \mathcal H\}$ instead of $\mathcal H\Delta\mathcal H$:
\begin{equation}
\mathrm{disc}_{h, \mathcal{H}}(P,Q) = \sup_{h'\in \mathcal H}
\left(\mathbb{E}_{ Q} L(h',h) - \mathbb{E}_{ P}L(h',h)\right).
\end{equation}
Based on these discrepancies, rigorous theories of generalization bounds for the target error were derived, which have inspired a line of influential works on domain adaptation \cite{cite:JMLR16RevGrad,cite:CVPR18MCD,cite:NIPS18CDAN,thy:zhang2019bridging}.

While making essential advances in domain adaptation theory, these discrepancies are all defined as the supremum over the \emph{whole} hypothesis space. 
This will include bad hypotheses impossible to have lower error and result in \emph{overestimation} of the generalization bound. 
Inspired by this observation, we introduce the localized hypothesis space and define the localized variants of these discrepancies:
\begin{definition}[\textbf{Localized Discrepancies}]
	For any source distribution $P$ and target distribution $Q$ on $\mathcal{X}\times \mathcal{Y}$, any hypothesis space $\mathcal{H}$ and any $r\ge 0$, the \emph{\textbf{localized hypothesis space}} $\mathcal{H}_r$ is defined as
	\begin{equation}\label{Hr}
	\mathcal H_r = \{h\in \mathcal{H}| \mathbb{E}_P L(h(x), y)\le r\}.
	\end{equation}
	Based on $\mathcal{H}_r$, the \emph{\textbf{localized $\mathcal{H}\Delta \mathcal{H}$-discrepancy}} from $P$ to $Q$ is defined as
	\begin{equation}\label{LHH}
	\mathrm{disc}_{\mathcal{H}_r} (P, Q)=\sup_{h, h'\in \mathcal H_r}
	\left(\mathbb{E}_{ Q} L(h',h) - \mathbb{E}_{ P}L(h',h)\right).
	\end{equation}
	And for any $h\in \mathcal{H}$, the \emph{\textbf{localized disparity discrepancy}} from $P$ to $Q$ is defined as
	\begin{equation}\label{LDD}
	\mathrm{disc}_{h, \mathcal{H}_r} (P, Q)=\sup_{h'\in \mathcal H_r}
	\left(\mathbb{E}_{ Q} L(h',h) - \mathbb{E}_{ P}L(h',h)\right).
	\end{equation}
\end{definition}

Note that $\mathrm{disc}_{\mathcal{H}_r} (P, Q)$ stresses the discrepancy used for \emph{one-way} transfer from $P$ to $Q$. The localized discrepancies are monotonically increasing functions for $0\le r\le 1$. Specifically, $\mathrm{disc}_{\mathcal{H}_r} (P, Q)\le \mathrm{disc}_{\mathcal{H}_1}(P, Q)=\mathrm{disc}_{\mathcal{H}}(P, Q)$ and $\mathrm{disc}_{h, \mathcal{H}_r} (P, Q)\le \mathrm{disc}_{h, \mathcal{H}_1} (P, Q)=\mathrm{disc}_{h, \mathcal{H}}(P,Q)$. 
Given a proper parameter $r$, we show that the localized discrepancies also imply target error bound.

\begin{theorem}[\textbf{Error Bound}]\label{UpperBound}
	For distributions $P$, $Q$ and hypothesis space $\mathcal{H}$, let $\lambda$ be the ideal joint error:
	\begin{equation}
	\lambda \triangleq \min_{h\in \mathcal{H}}\left( \err_{P}(h) + \err_{Q}(h) \right).
	\end{equation}
	Then for any $r> \lambda$ and $h\in \mathcal{H}_{r}$, we have the following upper bound of target error:
	\begin{equation} \label{eqn:ldbound}
	\err_{Q} (h) \le \err_P(h) +  \mathrm{disc}_{\mathcal{H}_r} (P, Q) + \lambda.
	\end{equation}
	Similarly, for any $r> \lambda$ and $h\in \mathcal{H}$, we have the following upper bound of target error:
	\begin{equation}
	\err_{Q} (h) \le \err_P(h) +  \mathrm{disc}_{h, \mathcal{H}_r} (P, Q) + \lambda.
	\end{equation}
\end{theorem}

These bounds share similar form as the seminal ones \cite{thy:shai10ad, thy:zhang2019bridging} except for using different discrepancies. The first term is source error. The ideal joint error $\lambda$ coincides with the adaptability term in \cite{thy:shai10ad}. Since these localized discrepancies are smaller, the resulting bounds are also relatively smaller than \cite{thy:shai10ad, thy:zhang2019bridging}.

Our bounds depend on the common \emph{assumption} that $\lambda$ is small \cite{thy:shai10ad}.
Previous works also show that a small $\lambda$ is necessary for the success of domain adaptation \cite{thy:shai2012hardness}.
This assumption actually implies a restricted space for the ideal hypothesis since it states that the source error of the ideal hypothesis is small.
Thus the target expected error can be bounded by the localized discrepancies under the mild \emph{assumption} $r> \lambda$. 
Although we cannot compute the accurate value of $\lambda$, we must assume that $\lambda$ is small such that the choice of $r$ could be more flexible; otherwise domain adaptation will fail.

\begin{remark}
	The idea of reducing the hypothesis space in discrepancy appears in previous works. 
	\cite{thy:cortes2015gd} proposes delicate generalized discrepancy when one of the hypothesis spaces in $\mathcal{H}\Delta\mathcal{H}$ is contracted to $\mathcal{H}''\subset \mathcal{H}$. 
	They choose $\mathcal{H}''$ with small amount of \emph{labeled target data} to ensure it is not far from the target ideal hypothesis. 
	\cite{thy:hanneke2019on} discusses the localization of discrepancy in hypothesis transfer setting when some \emph{labeled target data} is available.  And they do not provide detailed analysis.
	Furthermore, these works do not prove that the reduction of hypothesis space will lead to better generalization properties.
	\cite{thy:germain2013bayes} studies domain adaptation in PAC-Bayes paradigms. They consider discrepancy based on single hypothesis distribution. This will bring an incomputable term and need assumption that the term is small for all hypothesis distributions.
\end{remark}

\begin{remark}
	The localized discrepancies are no longer distance or even pseudo-metric on distributions. 
	First, we will show that $\mathrm{disc}_{\mathcal{H}_r} (P, Q)=0$ may not imply $P=Q$. 
	Second, we will show that the localized discrepancies may not be symmetric. 
	As the localized discrepancies have the ability to bound the target error, they are still \emph{proper-defined statistics} to measure the difference between two distributions.
\end{remark}

\begin{remark}
	In the upper bounds proposed in Theorem \ref{UpperBound}, the inequalities stand for $r> \lambda$. 
	Define the ideal joint hypothesis $h^* \triangleq {\arg\min}_{h\in \mathcal{H}}\left( \err_{P}(h) + \err_{Q}(h) \right)$.
	When $r=r_0<\lambda$, we may have $h^* \notin \mathcal{H}_{r}$. 
	At this time, the new $\lambda_{r_0}$ may increase when the discrepancy term $\mathrm{disc}_{\mathcal{H}_{r_0}} (P, Q)$ is smaller. 
	So there is a trade-off between the localized discrepancy and $\lambda$. For empirical use, choose a larger $r$ is safer, but may lead to a larger discrepancy estimation.
\end{remark}

\section{Comparison between Discrepancies} \label{sec4}

We will give several examples to show that the localized discrepancies lead to a relatively accurate estimation of the target expected error. 
This depends on two properties of the localized discrepancies. 
First, the localized discrepancies can be significantly smaller than previous discrepancies \cite{thy:shai10ad, thy:zhang2019bridging}. 
Second, the localized discrepancies lead to asymmetric estimations between two different distributions.
For clarity, we focus on the localized $\mathcal{H}\Delta\mathcal{H}$-discrepancy \eqref{LHH} in this section. In the next section, we will show that with specific objective function, these advantages of the localized $\mathcal{H}\Delta\mathcal{H}$-discrepancy can be shared by the localized disparity discrepancy \eqref{LDD}. 

\begin{figure*}[htbp]
	\centering
	\subfigure[Examples \ref{ex:41} \& \ref{ex:43} ]{
		\includegraphics[width=0.23\textwidth]{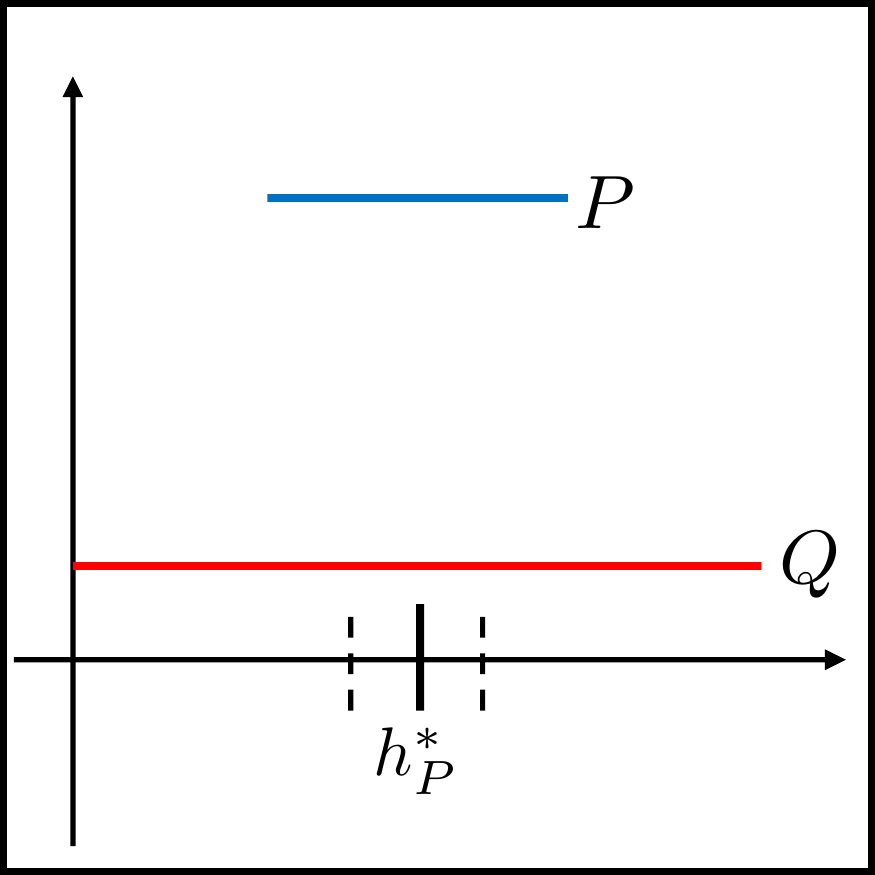}
		\label{fig:accuracy}
	}
	\hfil
	\subfigure[Example \ref{ex:42}]{
		\includegraphics[width=0.23\textwidth]{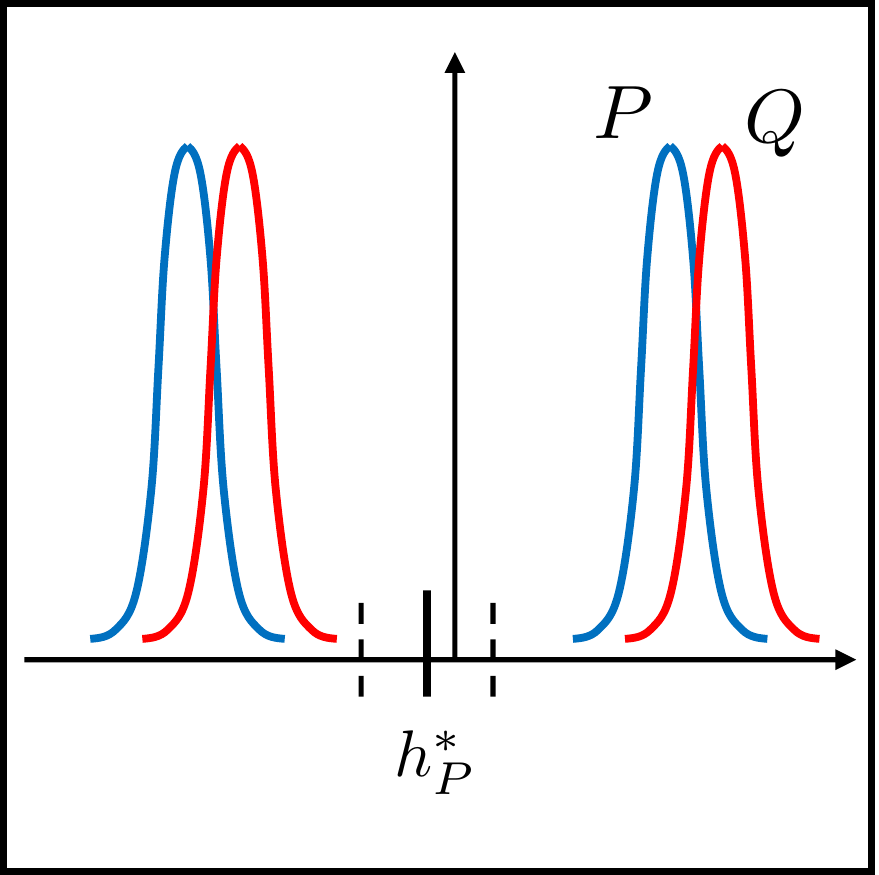}
		\label{fig:sourceeqvalue}
	}
	\hfil
	\subfigure[Example \ref{ex:44}]{
		\includegraphics[width=0.23\textwidth]{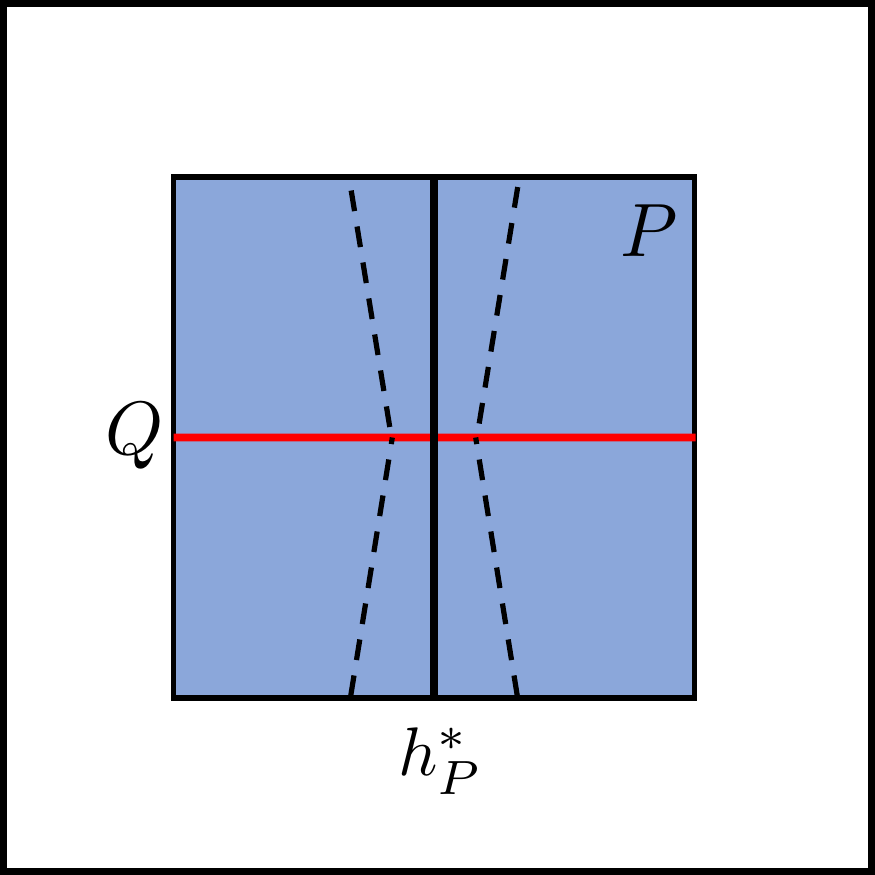}
		\label{fig:targeteqvalue}
	}
	\caption{Examples of Section~\ref{sec4}. The area in dash lines is the range of localized hypothesis space.}
	\label{fig:acc}
	\vspace{-10pt}
\end{figure*}

\subsection{Avoiding Overestimation}

In Section~\ref{sec3} we have shown that the localized $\mathcal{H}\Delta\mathcal{H}$-discrepancy is smaller than the $\mathcal{H}\Delta\mathcal{H}$-divergence \cite{thy:shai10ad}. We will further show that the localized $\mathcal{H}\Delta\mathcal{H}$-discrepancy can be strictly smaller.

\begin{example}[\textbf{Threshold Learning}] \label{ex:41}
	We first consider a commonly-used example. 
	Consider $\mathcal{X} = \mathbb{R}$ and $\mathcal{H}$ as the threshold function class.  
	A threshold function can be defined as $h_{t}$ that outputs $1$ on $x<t$ and $0$ otherwise. 
	The hypothesis space $\mathcal{H}$ contains $h_{t}$ and $1-h_{t}$ for all $ t\in \mathbb{R}$.
	At this time, the function $\mathbf{1}[h\ne h']$ in $\mathcal{H}\Delta\mathcal{H}$ can be written as $\mathbf{1}[x\in [t, t')]$, $\mathbf{1}[x\in (t, t']]$ or $\mathbf{1}[x\notin (t, t')]$ for some $t<t'$.
	We assume that the source distribution $Q$ is uniform distribution on $\mathcal{X}(Q)=[0,1]$ and the target distribution $P$ is uniform distribution on $\mathcal{X}(P) = [\frac{1}{2}-\epsilon,\frac{1}{2}+\epsilon]$.  
	The two domains share the same labeling function $h_{\frac{1}{2}}$.
	So it could be easily checked that $\mathrm{disc}_{\mathcal{H}} (P, Q)=1-2\epsilon$, $\mathrm{disc}_{h_{1/2}, \mathcal{H}}(P, Q)=\frac{1}{2}-\epsilon$ and $\mathrm{disc}_{\mathcal{H}_r} (P, Q)=0$ for all $r\in (0, \frac{1}{4})$. 
\end{example}

When $\epsilon$ is smaller, $\mathrm{disc}_{\mathcal{H}} (P, Q)$ will approach $1$. Even set $h=h_{1/2}=h^*$ as the ideal joint hypothesis, $\mathrm{disc}_{h, \mathcal{H}}(P, Q)$ is as large as $\frac{1}{2}$.
In this example, the localized discrepancy has significantly smaller value compared with the non-localized discrepancy.
This example shows that even the support sets of the two domains are highly non-overlapping, the domain adaptation is still possible. 
And this fact could be revealed by the localized discrepancy.

\begin{example}[\textbf{Gaussian Mixtures}] \label{ex:42}
	We then consider transfer between two Gaussian mixtures on $\mathcal{X} = \mathbb{R}$. The hypothesis space is the same as in Example \ref{ex:41}. We take $P=\frac{1}{2}\mathcal{N}(-10,1)+\frac{1}{2}\mathcal{N}(8,1)$ and $Q=\frac{1}{2}\mathcal{N}(-8,1)+\frac{1}{2}\mathcal{N}(10,1)$. We set source ideal hypothesis $h_P^{*}$ as $h_{-1}$ and target ideal hypothesis $h_Q^{*}$ as $h_{1}$. 
	At this time, the ideal joint hypothesis $h^{*} = h_{0}$.
	After simple calculations, we have $\mathrm{disc}_{\mathcal{H}} (P, Q)\ge 0.68$, $\mathrm{disc}_{{h_0}, \mathcal{H}}(P, Q)\ge 0.68$ and $\mathrm{disc}_{\mathcal{H}_r} (P, Q)<0.001$ for $r\in (\lambda, \sqrt{\lambda})$. 
\end{example} 

In this example, both distributions have good clustering structures. The inter-distance between centers of different class is about $18$.
And the intra-distance between source and target cluster centers in each class is $2$.
Although the intra-distance is not near since the source and the target do not closely overlap, the inter-distance is much larger and the transfer could also be safe.
The localized $\mathcal{H}\Delta\mathcal{H}$-discrepancy could capture this fact, when the other vanilla discrepancies fail to give satisfactory upper bounds.

\subsection{Asymmetric Properties}

We discuss the asymmetric properties of the localized discrepancies in this section.
\cite{cite:PAMI19DAN} finds that in experiments, the difficulties of transfer from $P$ to $Q$ can be significantly different with transfer from $Q$ to $P$. 
If we exchange $P$ and $Q$ in the previous discrepancies \cite{thy:shai10ad, thy:zhang2019bridging}, the upper bounds will remain the same.  
This contradiction indicates that we need a new discrepancy-based theory to address the asymmetric case. 
In this section, we will show that the localized discrepancies are able to reveal the asymmetric difficulties, thereby giving a preliminary answer to this open problem.

In the threshold learning problem (Example \ref{ex:41}), we can intuitively conjecture that the transfer from $P$ to $Q$ should be easier than from $Q$ to $P$, since $P$ is concentrated around the separating point.  
We now show that this fact could be revealed by the localized discrepancy.
\begin{example}[\textbf{Threshold Learning Rivisited}] \label{ex:43}
	We continue to discuss the threshold learning in Example \ref{ex:41}. When we take $P$ as target distribution and $Q$ as source distribution, the values of the three discrepancies change to  $\mathrm{disc}_{\mathcal{H}} (Q, P)=1-2\epsilon$, $\mathrm{disc}_{h_{\frac{1}{2}}, \mathcal{H}}(Q, P)=\frac{1}{2}-\epsilon$ and $\mathrm{disc}_{\mathcal{H}}^{r} (Q, P)=r(\frac{1}{\epsilon}-2)$ for all $r\in(0, \epsilon]$.
\end{example}

For $r(\frac{1}{\epsilon}-2)\ge 0$, the localized $\mathcal{H}\Delta\mathcal{H}$-discrepancy can capture the asymmetric difficulties while other discrepancies cannot. 
The localized $\mathcal{H}\Delta\mathcal{H}$-discrepancy also shows that under a stronger restriction on parameter $r$, the discrepancy could be smaller and the transfer could be easier. 

\begin{example}[\textbf{Target on Low-dimensional Manifold}] \label{ex:44}
	Consider $\mathcal{X} = [0,1]^2$ and $\mathcal{H}$ as the linear classifier space. 
	Let source $P$ be uniform distribution on $ [0,1]^2$ and target $Q$ be uniform distribution on line segment $\{(t,\frac{1}{2})|0\le t\le 1\}$. 
	Set the labeling function $l$ as $l(x_1, x_2) = \mathbf{1}[x_1 > \frac{1}{2}]$.
	We can prove that $\mathrm{disc}_{\mathcal{H}} (P, Q)=1$, $\mathrm{disc}_{l, \mathcal{H}}(P, Q)\ge\frac{1}{2}$ and $\mathrm{disc}_{\mathcal{H}_r} (P, Q)=0$ for all $r\in(0,\frac{1}{4})$.
	After exchanging the source and target domains, we have $\mathrm{disc}_{\mathcal{H}} (Q, P)=\mathrm{disc}_{\mathcal{H}_r} (Q, P)=1$. At this time, it is impossible to learn the ideal hypothesis $l$ with only source data from $Q$ and the disparity discrepancy will also be large. 
\end{example}

In this example, target domain lies on a low-dimensional manifold of source domain support set. Transfer from low-dimensional $Q$ to high-dimensional $P$ is almost impossible.
As previous work \cite{thy:samory18marginal} illustrates, it is hard for discrepancy-based theories to cope with this setting. 
If the discrepancy is symmetric, then the discrepancy on transfer from $P$ to $Q$ will be as large as transfer from $Q$ to $P$. 
We show that the localized $\mathcal{H}\Delta\mathcal{H}$-discrepancy is asymmetric and could be very small when transfer from $P$ to $Q$. 
Thus the localized discrepancies could provide theoretical foundations for this setting.

\section{Improving Generalization Bounds with Localized Discrepancies} \label{sec5}

In this section, we propose novel generalization bounds for domain adaptation based on the localized discrepancies. 
In the seminal theories \cite{thy:shai10ad,cite:COLT09DAT,thy:zhang2019bridging}, the sample complexity terms of the generalization bounds are commonly proved as $O(\sqrt{d\log n/n}+\sqrt{d\log m/m})$, where $d$ is VC dimension of the hypothesis space $\mathcal{H}$. 
We prove that the localized discrepancies could lead to improved generalization bounds, and a notable new finding is that the sample complexity terms depend on the value of the localized discrepancies. 
We will first give bounds based on the localized $\mathcal{H}\Delta\mathcal{H}$-discrepancy \eqref{LHH} and then bounds on the localized disparity discrepancy \eqref{LDD}.

\subsection{Generalization Bound with Localized $\mathcal{H}\Delta\mathcal{H}$-Discrepancy}

We begin with the binary classification setting using 0-1 loss. 
We first introduce an objective function for estimating and minimizing the target error bound based on datasets $P_n$ and $Q_m$.
Recall from Theorem \ref{UpperBound} that the target error bound based on the localized $\mathcal{H}\Delta\mathcal{H}$-discrepancy is for all $h\in \mathcal{H}_r$,
\begin{equation}
\err_{Q}(h)\le \err_{P}(h) +  \mathrm{disc}_{\mathcal{H}_r} (P, Q) +\lambda.
\end{equation}
We adopt the common assumption that $\lambda$ is small \cite{thy:shai10ad}.
As the computations of the first two terms are separated, we can first minimize the error of hypothesis $h$ on source labeled data and then estimate the localized $\mathcal{H}\Delta\mathcal{H}$-discrepancy on both source and target unlabeled data. 
Recall that Theorem \ref{UpperBound} requires the main hypothesis to satisfy $h\in \mathcal{H}_r$.  
However, this cannot be directly examined since the original distributions are unknown. 
To solve this problem, we show that a localized hypothesis space $\tilde{\mathcal{H}}_{r^-}$ with small error on the source dataset can be included in $\mathcal{H}_{r}$ with high probability.
Estimation of the localized $\mathcal{H}\Delta\mathcal{H}$-discrepancy can be seen as a constrained optimization problem:
\begin{equation}
\sup_{h,h'\in \mathcal{H}_r}{(\mathbb{E}_{ Q_m} L(h',h) - \mathbb{E}_{ P_n}L(h',h))}.
\end{equation}
Again, the problem of deciding whether hypothesis $h$ is in $\mathcal{H}_r$ is unsolvable. 
To address this difficulty, we prove that a localized hypothesis space $\tilde{\mathcal{H}}_{r^+}$ with small error on the source dataset can subsume $\mathcal{H}_{r}$ with high probability. The definitions of the two localized hypothesis spaces are given below.

\begin{definition}\label{def:5.1}
	For distribution $P$ and its sample $P_n$, binary classifier space $\mathcal{H}$ with VC dimension $d$, and any $\delta>0$, let $\mathcal{E}= 4\frac{d\left(1+4\ln (n/d)\right)-\ln \delta / 16}{n} $. For all $r>\mathcal{E}$,  set $\mathfrak{C}^{+}(n,d, \delta, r)=\frac{\mathcal{E}}{2}(1+\sqrt{1+\frac{4 r}{\mathcal{E}}})$, $\mathfrak{C}^{-}(n,d, \delta, r) = \sqrt{\mathcal{E}r}$ and define the \emph{localized hypothesis spaces} $\tilde{\mathcal{H}}_{r^+}$, $\tilde{\mathcal{H}}_{r^-}$ as
	\begin{equation}
	\begin{aligned}
	\tilde{\mathcal{H}}_{r^+}= \{h\in \mathcal{H}| \mathbb{E}_{P_n} L(h(x), y)\le r + \mathfrak{C}^{+}(n,d, \delta, r)\},
	\\
	\tilde{\mathcal{H}}_{r^-}= \{h\in \mathcal{H}| \mathbb{E}_{P_n} L(h(x), y)\le r - \mathfrak{C}^{-}(n,d, \delta, r)\}.
	\end{aligned}
	\end{equation}
\end{definition}

\begin{lemma}\label{lem:5.2}
	With probability no less than $1-\frac{\delta}{2}$, we have
	$\tilde{\mathcal{H}}_{r^-} \subset \mathcal{H}_{r}\subset \tilde{\mathcal{H}}_{r^+}$.
\end{lemma}

Thus the objective for finding hypothesis $h$ that generalizes across domains can be written as
\begin{equation}\label{obj:1}
\min_{h\in \tilde{\mathcal{H}}_{r^-}}\{\err_{P_n}(h)\} +  \mathrm{disc}_{\tilde{\mathcal{H}}_{r^+}} (P_n,Q_m).
\end{equation}
Note that the constraint $h\in \tilde{\mathcal{H}}_{r^-}$ could be naturally satisfied for $r>\lambda$ and large enough $n$.
With this objective, we derive the following generalization bound for binary classification.

\begin{theorem}[\textbf{Generalization Bound with Localized $\mathcal{H}\Delta\mathcal{H}$-Discrepancy}]\label{thm:5.3}
	For distributions $P$, $Q$ on $\mathcal{X}\times \mathcal{Y}$, their empirical distributions $P_n$, $Q_m$, binary classifier space $\mathcal{H}$ with VC dimension $d$ and $\delta > 0$, let $\lambda$ be the ideal joint error and set $\mathcal{E}$ as Definition \ref{def:5.1}. Set fixed $r> \mathcal{E} + \lambda$.  Let $\hat h$ be the solution of objective \eqref{obj:1}. Then with probability no less than $1-\delta$, we have
	\begin{equation}
	\begin{aligned}
	\err_{Q} (\hat h) & \le  \err_{P_n}(\hat h) +  \mathrm{disc}_{\tilde{\mathcal{H}}_{r^+}} (P_n, Q_m)+ \lambda + O( \frac{d\log n +\log(1/\delta)}{n}) + O( \frac{d\log m +\log(1/\delta)}{m}) 
	\\
	&+ O( \sqrt{ \frac{2r (d\log n +\log(1/\delta))}{n}}) + O( \sqrt{ \frac{(\mathrm{disc}_{\tilde{\mathcal{H}}_{r^+}} (P_n, Q_m)+2r)(d\log m+\log(1/\delta)) }{m}}).
	\end{aligned}
	\end{equation}
\end{theorem}

This bound shows that the localized discrepancy also influences the generalization error.
The localized parameter $r$ plays a dominant role in controlling the source sample complexity, while both $r$ and the empirical localized $\mathcal{H}\Delta\mathcal{H}$-discrepancy control the target complexity. 
When these terms are smaller, the rate of this upper bound can be faster than previous bounds with $O(\sqrt{d\log n/n}+\sqrt{d\log m/m})$ complexity.
When $P \approx Q$ and $r\approx \lambda$, the complexity term has similar order with classic supervised learning bound $O(\sqrt{\err_{D}(h)d\log n/n} +d\log n/n)$ \cite{vapnik1998statistical, thy:boucheron2005theory}.
Except the negligible $\lambda$ term, the right hand of the generalization bound can be empirically computed from finite samples. 
With a proper choice of $r$, the upper bound may provide an explicit guidance on algorithm designs.

The main idea behind the proof (deferred to the appendix) is that the localized $\mathcal{H}\Delta\mathcal{H}$-discrepancy could control the function class that is reachable during estimation.
Previous discrepancies such as $\mathcal{H}\Delta\mathcal{H}$-divergence \cite{thy:shai10ad} takes the supremum on the \emph{whole} hypothesis space, while the supremum may be reached by any hypothesis in the whole space. This fact limits the generalization ability of these discrepancies.
The localized $\mathcal{H}\Delta\mathcal{H}$-discrepancy excludes a large portion of unreachable hypotheses and yields remarkable reduction of both discrepancy value and generalization error.


\subsection{Generalization Bound with Localized Disparity Discrepancy}

We further derive generalization bound with the localized disparity discrepancy.
Similar with the objective \eqref{obj:1}, the localized disparity discrepancy leads to another objective on datasets $P_n$ and $Q_m$:
\begin{equation}\label{obj:2}
\min_{h\in \mathcal{H}}\{\err_{P_n}(h) +  \mathrm{disc}_{h,\tilde{\mathcal{H}}_{r^+}} (P_n,Q_m)\} \triangleq \min_{h\in \mathcal{H}} \mathfrak{O}(h, \mathcal{H}, r^{+}, P_n, Q_m).
\end{equation}
The localized hypothesis space $\tilde{\mathcal{H}}_{r^{+}} $ follows Definition \ref{def:5.1}.  
The upper bound led by the localized disparity discrepancy is more flexible and it does not require the main hypothesis to satisfy $h\in \mathcal{H}_{r^{-}}$. 
It is worthy to note that the two terms in the objective \eqref{obj:2} are no longer separated. 
The main classifier $h$ also participates in the estimation of the discrepancy.  
Compared with objective \eqref{obj:1}, the solution of objective \eqref{obj:2} leads to a smaller upper bound. This is explained in the following proposition.

\begin{proposition}\label{pro:lddbetter}
	Set $\hat h$ as the solution of objective \eqref{obj:1} and $\check{h}$ as the solution of objective \eqref{obj:2}. We have the following results:
	\begin{equation}
	\mathfrak{O}(\check{h}, \mathcal{H}, r^{+}, P_n, Q_m)  \le \err_{P_n}(\hat{h}) +  \mathrm{disc}_{\tilde{\mathcal{H}}_{r^+}} (P_n,Q_m) \le r +  \mathrm{disc}_{\tilde{\mathcal{H}}_{r^+}} (P_n,Q_m) .
	\end{equation}
\end{proposition}

As the definition of the discrepancy is hypothesis-dependent and more flexible, the localized disparity discrepancy can further avoid \emph{overestimation}.
Under the objective \eqref{obj:2}, the upper bound induced by the localized disparity discrepancy is given in the following theorem.

\begin{theorem}[\textbf{Generalization Bound with Localized Disparity Discrepancy}]
	For distributions $P$, $Q$ on $\mathcal{X}\times \mathcal{Y}$, their empirical distributions $P_n$, $Q_m$, binary classifier space $\mathcal{H}$ and $\delta > 0$, set fixed $r> \lambda$.  Let $\check h$ be the solution of objective \eqref{obj:2}. Then with probability no less than $1-\delta$, we have
	\begin{equation}
	\begin{aligned}
	&\err_{Q} (\check h) \le  
	\err_{P_n}( \check h) +  
	\mathrm{disc}_{\check{h}, \tilde{\mathcal{H}}_{r^+}} (P_n, Q_m)
	+ \lambda + O( \frac{d\log n +\log(1/\delta)}{n}) + 
	O( \frac{d\log m +\log(1/\delta) }{m}) 
	\\
	&+ O( \sqrt{ \frac{(\err_{P_n}( \check h)+r )(d\log n+\log(1/\delta)) }{n}}) + O( \sqrt{ \frac{(\mathfrak{O}(\check h, \mathcal{H}, r^{+}, P_n, Q_m)+r)(d\log m+\log(1/\delta)) }{m}}).
	\end{aligned}
	\end{equation}
\end{theorem}
This bound inherits the advantages of the bound proposed in Theorem \ref{thm:5.3}.
Moreover, this bound is slightly improved when $\err_{P_n}( \check h)<r$ due to the fact proposed in Proposition \ref{pro:lddbetter}. 

\section{Super Transfer with Localized Discrepancies} \label{sec6}
Finally, we delve into the asymmetric sample complexity with the localized $\mathcal{H}\Delta\mathcal{H}$-discrepancy.
Hanneke \textit{et al.} \cite{thy:hanneke2019on} initiate ``super transfer'' in hypothesis transfer setting to describe that source domain enjoys a \emph{faster} sample complexity compared with target domain.
In the previous bounds, we can see that both source and target sample complexities share the same order. 
To achieve super transfer in unsupervised domain adaptation, we further extend the localized discrepancies to a \emph{boosted} version. 
Inspired by the ``transfer exponent'' proposed in \cite{thy:hanneke2019on}, we define the $\gamma$-boosted localized $\mathcal{H}\Delta\mathcal{H}$-discrepancy and derive the target error upper bound as follows:

\begin{definition}
	For distributions $P$, $Q$ on $\mathcal{X}\times \mathcal{Y}$, hypothesis space $\mathcal{H}$ and $r>0$, the \emph{$\gamma$\textbf{-boosted localized $\mathcal{H}\Delta\mathcal{H}$-discrepancy}} is defined as
	\begin{equation}
	\mathrm{disc}_{\mathcal{H}_{r}}^{\gamma} (P, Q)=\sup_{h, h'\in \mathcal H_r}
	\left(\mathbb{E}_{ Q} L(h',h) - (\mathbb{E}_{ P}L(h',h))^\gamma\right).
	\end{equation}
\end{definition}

\begin{theorem} 
	For distributions $P$, $Q$ on $\mathcal{X}\times \mathcal{Y}$ and  hypothesis space $\mathcal{H}$, let $\lambda$ be the ideal joint error. Assume $\lambda < \frac{1}{2}$. Then for any $\gamma \ge 1$, $r \in (\lambda,\frac{1}{2})$ and $h\in \mathcal{H}_{r}$, we have the following upper bound of target error,
	\begin{equation} \label{eqn:superbound}
	\err_{Q} (h) \le 2^{\gamma - 1}(\err_P(h))^{\gamma} +  \mathrm{disc}_{\mathcal{H}_{r}}^{\gamma} (P, Q) +    \lambda.
	\end{equation}
\end{theorem}

When $\gamma = 1$, the upper bound \eqref{eqn:superbound} coincides with the bound with the localized $\mathcal{H}\Delta\mathcal{H}$-discrepancy \eqref{eqn:ldbound}.
As $\err_P(h) < \frac{1}{2}$ for all $h\in \mathcal{H}_{r}$, we have $2^{\gamma - 1}(\err_P(h))^{\gamma} = (2\err_P(h))^{\gamma - 1}\err_P(h) < \err_P(h)$ for all $\gamma > 1$. 
Although the coefficient is $2^{\gamma - 1}$, the source error term is actually smaller for all $\gamma > 1$. 
This bound shows interesting trade-off between discrepancy and the source error term. 
When $\gamma$ grows larger, the source error term will be smaller and the discrepancy will be larger. 
After introducing the upper bound led by the $\gamma$-boosted localized discrepancy, we study the generalization properties. Akin to Equation \eqref{obj:1}, we define the objective led by the $\gamma$-boosted localized $\mathcal{H}\Delta\mathcal{H}$-discrepancy as
\begin{equation}\label{obj:3}
\min_{h\in \tilde{\mathcal{H}}_{r^-}}\{\err_{P_n}(h)\} +  \mathrm{disc}_{\tilde{\mathcal{H}}_{r^+}}^{\gamma} (P_n,Q_m),
\end{equation}
where the localized hypothesis spaces $\tilde{\mathcal{H}}_{r^+}$ and $\tilde{\mathcal{H}}_{r^-}$ follow Definition \ref{def:5.1}.

\begin{theorem}[\textbf{Generalization Bound with $\gamma$-Boosted Localized $\mathcal{H}\Delta\mathcal{H}$-Discrepancy}]
	For distributions $P$, $Q$ on $\mathcal{X}\times \mathcal{Y}$, their empirical distributions $P_n$, $Q_m$, binary classifier space $\mathcal{H}$ with VC dimension $d$, $\gamma \ge 1$ and $\delta > 0$, let $\lambda$ be the ideal joint error and set $\mathcal{E}$ as Definition \ref{def:5.1}. Set fixed $r> \mathcal{E} + \lambda$.  Let $\hat h$ be the solution of objective \eqref{obj:1}. Then with probability no less than $1-\delta$, we have
	\begin{equation}
	\begin{aligned}
	&\err_{Q} (\hat h) \le \err_{P_n}(\hat h) +  \mathrm{disc}_{\tilde{\mathcal{H}}_{r^+}} (P_n, Q_m)+ \lambda +( O( \frac{d\log n +\log(1/\delta)}{n}) )^\gamma+ O( \frac{d\log m +\log(1/\delta)}{m})
	\\
	&+ (O( \sqrt{ \frac{2r (d\log n+\log(1/\delta)) }{n}}) )^\gamma + O( \sqrt{ \frac{(\mathrm{disc}_{\tilde{\mathcal{H}}_{r^+}} (P_n, Q_m)+(2r)^\gamma)(d\log m+\log(1/\delta)) }{m}}).
	\end{aligned}
	\end{equation}
\end{theorem}
With the $\gamma$-boosted localized $\mathcal{H}\Delta\mathcal{H}$-discrepancy, the rates of source and target sample complexities become different. 
The rate of the source domain can be faster and more sample-efficient than that of the target domain, thus the super transfer can be achieved. 
We will discuss the bounds for general loss family, the $\gamma$-boosted localized disparity discrepancy and the setting of $0<\gamma<1$ in appendix.

\section{Related Work}

\paragraph{Theory for Domain Adaptation}
One of the mainstream domain adaptation theoretical frameworks is the discrepancy-based theory. 
The seminal work \cite{cite:NIPS07DAT} first proposes $\mathcal{H}\Delta\mathcal{H}$-divergence for binary classification. \cite{cite:COLT09DAT} extends the theory to a wider class of tasks and gives further theoretical analysis for regularized convex problems.
After that, \cite{thy:cortes2014reg} proposes detailed studies on domain adaptation in regression tasks.
\cite{thy:cortes2015gd} proposes delicate generalized discrepancy which shares the advantages of discrepancy-based method and importance weighting.
\cite{thy:zhang2019bridging} comes up with the disparity discrepancy which is hypothesis-dependent. 
\cite{thy:germain2013bayes} studies domain adaptation problem with discrepancy from the PAC-Bayesian perspective.
\cite{thy:mohri2012ydsic} adapts the discrepancy to domain adaptation problem with available target labeled data.

Another line of theories focus on adopting classical distribution distances in domain adaptation.
\cite{thy:courty2017OT, thy:redko2017otda, cite:NIPS17JDOT} consider using Wasserstein distance as surrogate of the parameterized discrepancy.
\cite{cite:ICML15DAN, thy:redko2016mmd} adopt Kernel Maximum Mean Discrepancy \cite{cite:JMLR12MMD} as an upper bound of the previous discrepancies.
In addition, \cite{thy:germain2016bayes, thy:johansson2019support} derive upper bounds for target error with divergences defined with the density ratio.
There are also works concentrating on domain adaptation with importance weighting methods \cite{cite:JMLR07IWCV, thy:sugiyama2008direct, thy:cortes2010importanceweighting}.
Domain adaptation problem is analyzed as structured generalization models from the causal view \cite{thy:zhang2013causal, thy:gong2016ctc}. 
A series of works \cite{thy:shai2014da, thy:scott2018a,thy:samory18marginal} provide theoretical results for domain adaptation with non-parametric models.

\paragraph{Theory for Transfer Learning}
There are also other paradigms in the broader transfer learning area.
\cite{thy:kuzborskij2013stability, thy:perrot2015metric, thy:du2017hypothesis, thy:hanneke2019on} study hypothesis transfer which learns from smaller target labeled data and auxiliary hypothesis.
\cite{thy:yang2013active, thy:chattopadhyay2013active, thy:berlind2015active} study active transfer learning which allows requirements for a few labels of target data.
Multitask learning problem concentrates on how to share knowledge between different tasks for better overall results 
\cite{thy:maurer2006MTL,thy:pentina2015multiandlifelong,thy:maurer2016benefit,thy:pentina2017multi,thy:wu2020understanding}.
Lifelong learning or learning to learn maximizes the overall performance on the tasks drawn from underlying task distribution \cite{thy:pentina2014bayes, thy:pentina2015lifelong,thy:khodak2019provable}.

\section{Conclusion}
In this paper, we define several localized discrepancies and study them from multiple perspectives. 
We show that the localized discrepancies can significantly avoid overestimation and reveal asymmetric transfer difficulties. 
We prove improved generalization bounds, and extend the localized discrepancies to the boosted versions, leading to a generalization bound that achieves super transfer.

\vskip 0.2in
\bibliography{paper}
\end{document}